# Overview of Stemming Algorithms for Indian and Non-Indian Languages

Dalwadi Bijal [#1], Suthar Sanket [*2]

[#1] *Department of Information Technology,*
*BVM Engineering College,Vallabh Vidya Nagar, India*

[#2] *Department of Information Technology,*
*Charusat University, Changa, India*

*Abstract*— **Stemming is a pre-processing step in Text Mining applications as well as a very common requirement of Natural Language processing functions. Stemming is the process for reducing inflected words to their stem. The main purpose of stemming is to reduce different grammatical forms / word forms of a word like its noun, adjective, verb, adverb etc. to its root form. Stemming is widely uses in Information Retrieval system and reduces the size of index files. We can say that the goal of stemming is to reduce inflectional forms and sometimes derivationally related forms of a word to a common base form. In this paper we have discussed different stemming algorithm for non-Indian and Indian language, methods of stemming, accuracy and errors.**

*Keywords*— **Over-stemming, Under-stemming, Rule based stemming.**

## I. INTRODUCTION

Many natural languages like Dravidian languages (Tamil, Telugu, Malayalam and Kannada), Finno-Ugric languages (Finnish, Estonian, Hungarian, Turkish), Indo-Aryan languages (Hindi, Bengali, Marathi, Gujarati) are inflected. In such languages several words sharing the same morphological invariant (root) can be related to the same topic. The ability of an Information Retrieval (IR) system to conflate words allows reducing index and enhancing recall. In most cases, morphological variants of words have similar semantic interpretations and can be considered as equivalent for the purpose of IR applications. For this reason, a no of stemmers have been developed, which attempt to reduce a word to its stem or root form. Each algorithm attempts to convert the morphological variants of a word like 'playing', 'played', 'plays' etc. to get mapped to the word 'play'. Thus, the key terms of a query or document are represented by stems rather than by the original words.

## II. STEMMING ALGORITHMS FOR INDIAN LANGUAGE

Many Indian languages are in highly inflected form. Indian languages are Tamil, Malayalam, Gujarati, Marathi, Hindi, Bengali etc.

### A. Assamese stemming
Navanath Saharia, Utpal Sharma and Jugal Kalita [6], they adopt suffix stripping approach along with a rule engine that generates all possible suffix sequences. They used two approaches. In first approach, they manually collected all possible suffixes and describe the suffix stripping approach.

### B. Gujarati Stemming
Pratikkumar Patel Kashyap Popat and Pushpak Bhattacharyya [1], they have used the EMILLE corpus for training and hand-crafted suffixes. Their approach is based on Goldsmith's take-all-splits method. They try to obtain the optimal split position for each word present in the Gujarati word list provided for training.
Kartik Suba, Dipti Jiandani and Pushpak Bhattacharyya [4], they present two stemmers for Gujarati – a lightweight inflectional stemmer based on hybrid approach and a heavyweight derivational stemmer based on a rule-based approach. For inflectional stemmer, they used POS based stemming and suffix stripping based on linguistic rules and used take-all-spilt method to obtain optimal split position of words. For derivational stemmer, they used suffix stripping, substitution and orthographic rules.
Juhi Ameta, Nisheeth Joshi, Iti Mathur [9], they present implementation of rule based stemmer of Gujarati. They have shown the creation rule for stemming and verifying it with human expert. The process of this stemmer is strips the suffixes based on the longest match.

### C. Hindi Stemming
Ananthakrishnan Ramanathan and Durgesh D Rao [8], present a lightweight stemmer for Hindi, which conflates terms by suffix removal. The suffix list was developed. The stemmer is implemented by simply removing from each word the longest possible suffix from suffix list. They have evaluated our stemmer by computing the number of under-stemming and over-stemming errors for a corpus of documents.
Upendra Mishra and Chandra Prakash [2], present the Hybrid approach which is combination of brute force and suffix removal approach and reduces the problem of over-stemming and under-stemming.

TABLE 1 ANALYSIS OF STEMMING ALGORITHM FOR INDIAN LANGUAGE

| Proposed Methods | Approach used | Year of Methods | Name of authors | Tested on Language | Dataset | Total words | Accuracy | Errors |
|---|---|---|---|---|---|---|---|---|
| Longest matched | Rule based | 2003 | Ananthakrishnan Ramanathan , Durgesh D Rao | Hindi | Online Hindi news, magazine, films, health, business, sports & politics. | 35977 | 88% | Overstemming -13.84% Understemming - 4.68% |
| Take-all-split method | Hand-crafted suffixes | 2010 | Pratikkumar Patel, Kashyap Popat | Gujarati | EMILLE corpus | Not mentioned | 67.86% | Not mentioned |
| FSA (Finite State Automata) | Morphotatic ruels | 2010 | Vijay Sundar Ram R, Sobha Lalitha Devi | Malayalam | Online Malayalam newspaper, mathrubhumi | Not mentioned | 94.76% | Not mentioned |
| n-gram method | Suffix stripping | 2010 | Mudassar M. Majgaonker ; Tanveer J Siddiqui | Marathi | Marathi corpus, internet | 132895 | 82.5% | Not mentioned |
| Take-all-split , POS based | Suffix stripping, Rule based | 2011 | Kartik Suba, Dipti Jiandani | Gujarati | EMILLE corpus | Not mentioned | 90.7% | Not mentioned |
| Brute force technique | Suffix Stripping | 2011 | Dinesh Kumar, Prince Rana | Punjabi | Online newspapers, dictionaries, articles | 52000 | 81.27% | Not mentioned |
| Longest matched | Rule based | 2012 | Juhi Ameta, Nisheeth Joshi, Iti Mathur | Gujarati | EMILLE corpus | 3000 | 91.5% | Not mentioned |
| Brute Force technique | Suffix stripping | 2012 | Upendra Mishra, Chandra Prakash | Hindi | | 15000 | 91.59% | Not mentioned |
| Lookup method | Suffix stripping | 2012 | Navanath Saharia | Assamese | EMILEE corpus | 123753 | 82% | Not mentioned |

### D. Punjabi stemming

Dinesh Kumar and Prince Rana proposed algorithm for Punjabi language. They used brute force approach. This approach employs a lookup table which contains relations between root words and inflected words. Brute force requires immense amount of storage to create a database but it reduces the problem of under-stemming and over-stemming. They have used suffix stripping if the word is not found in the database.

## III. STEMMING ALGORITHM FOR NON-INDIAN LANGUAGE

### A. Arabic stemming

Haidar Harmanani et al's [5] proposed an extensible method for natural languages indexing and search. The method is based on a rule engine that allows the system to be adapted to a variety of natural language without the need to develop a specialized IR system. The proposed method used full-text indexing and ranks terms according to their degree of relevance.

### B. English stemming

The first popular and effective stemmer for English is proposed by Lovins in 1968. The Lovins stemmer is a single pass stemmer, context sensitive and longest match stemmer. Lovins stemmer maintains a list of most frequent suffixes, and removes the longest suffix.

Porters stemming algorithm is as of now one of the most popular stemming methods. It is based on the idea that the suffixes in the English language (approximately 1200) are mostly made up of a combination of smaller and simpler suffixes. It has five steps, and within each step, rules are applied until one of them passes the conditions. The rule looks like the following:

<condition> <suffix> → <new suffix>

### C. Farsi / Persian stemming

The Farsi stemmer uses a Deterministic Finite Automata (DFA). The DFA is input string is obtained by reversing the stemmer is input string. It matches words with a set of suffixes and use multiple phases conforming to the rules of suffix stacking.

.

TABLE 2 ANALYSIS OF STEMMING ALGORITHM FOR NON-INDIAN LANGUAGE

| Proposed Methods | Approach used | Year of Methods | Name of author | Tested on Language | Dataset | Total words | Accuracy |
|---|---|---|---|---|---|---|---|
| Method proposed by Paice | Suffix stripping , Rule based | 2001 | Orengo, V.M. | Portuguese | Online newspapers | 2800 | 96% |
| Longest possible | Suffix stripping | 2003 | Preslav Nakov | Bulgarian | Morphological dictionary of Bulgarian | 889665 | Not mentioned |
| DFA (Deterministic Finite Automata) | Suffix stripping | 2003 | Kazem Taghva, Russell Beckley | Farsi / Persian | 1647 Farsi document, internet document | Not mentioned | Not mentioned |
| Stem based method on rule engine | Rule based | 2006 | Haidar Harmanani, Walid keirouz | Arabic | Online Arabic documents | Not mentioned | 100% |
| Suffix stripping | Derivational Suffix stripping | 2006 | Jacques Savoy | French | Online newspapers | Not mentioned | Not mentioned |

### D. Portuguese stemming

Orengo,V.M. describes the development of a simple and effective suffix-stripping algorithm for Portuguese. Each rule specifies the suffix to be removed; the minimum length allowed for the stem; a replacement suffix, if necessary and a list of exceptions. The longest possible suffix is always removed first because of the order of the rules within a step. It is based on a set of steps composed by a collection of rules

## IV. CONCLUSION

We have presented a comparative study of various stemming algorithms for Indian and non-Indian language. In this we studied that stemming significantly increases the retrieval results for both rule based and statistical approach. It is also useful in reducing the size of index files as the number of words to be indexed are reduced to common forms or so called stems. Most of stemming algorithms are based on rule based approach. The performance of rule based stemmer is superior to some well known method like brute force. Dictionary-based algorithms, including natural language processing approaches, allow integration with other applications, for example, for interactive query expansion or machine translation. Of course, they require constant updates to dictionaries due to language evolution, but this task is constantly performed by publishers and scientific groups. Also, growing computers power makes natural language processing approaches more feasible.